# Detecting Damage Building Using Real-time Crowdsourced Images and Transfer Learning

## Author Information

Gaurav Chachra

**Zymergen**

Qingkai Kong

**Lawrence Livermore National Laboratory & Berkeley Seismological Laboratory**

Jim Huang

**AT&T**

Srujay Korlakunta

**University of California, Berkeley**

Jennifer Grannen

**University of California, Berkeley**

Alexander Robson

**Berkeley Seismological Laboratory**

Richard Allen

**Berkeley Seismological Laboratory**

## Contributions

G.C. contributed to the design of the experiment, collection and labelling of the data, and conducted the experiment. Q.K. contributed to the design of the experiment, labelling the data and writing of the manuscript. J.H contributed to the labelling of the data and discussion of the solutions to the problems. S.K. contributed to the data labelling and implementation of the work pipeline. J.G. contributed to the data labelling and implementation of the work pipeline. A.R. contributed to the data collection and data labelling. R.A. contributed to the MyShake project and discussion of the solutions to the problems posed in this manuscript.

## Corresponding author

Correspondence to: Qingkai Kong

## Abstract

After significant earthquakes, we can see images posted on social media platforms by individuals and media agencies owing to the mass usage of smartphones these days. These images can be utilized to provide information about the shaking damage in the earthquake region both to the public and research community, and potentially to guide rescue work. This paper presents an automated way to extract the damaged building images after earthquakes from social media platforms such as Twitter and thus identify the particular user posts containing such images. Using transfer learning and ~6500 manually labelled images, we trained a deep learning model to recognize images with damaged buildings in the scene. The trained model achieved good performance when tested on newly acquired images of earthquakes at different locations and ran in near real-time on Twitter feed after the 2020 M7.0 earthquake in Turkey. Furthermore, to better understand how the model makes decisions, we also implemented the Grad-CAM method to visualize the important locations on the images that facilitate the decision.

## Introduction

Large earthquakes near populated areas leave unforgettable testimony of their intimidating powers: collapsed buildings, offset roads, disrupted lives, etc. With the help of current technology, images of this damage can be captured soon after the earthquake and uploaded online so that people in other parts of the world can easily browse through these images and videos to obtain a sense of the consequences of the disaster. Such images or videos taken by individuals or news media often contain important information about the damage, such as damaged buildings, bridges, roads and other infrastructures (Alam, Imran, and Ofli 2017; "Social Media in Disaster Japan" 2012).

There are existing efforts to extract useful information from these crowdsourcing images to better understand natural disasters. (Nguyen et al. 2017) used a transfer learning approach, more specifically, domain-specific fine-tuning Convolution Neural Network (CNN) to estimate the damage severity in the images on social media platforms after a disaster. (Alam, Ofli, and Imran 2018) provided a social media image processing pipeline to combine human and machine intelligence to extract information from images as a situational awareness task during a crisis event, such as earthquakes or hurricanes. They used the pre-trained VGG-16 model with fine-tuning to filter the relevant images from the collected images on Twitter. After the removal of the duplicated images and crowdsourcing efforts to label the images, they feed these images into an image classifier to identify damaged structures, injured people and rescue efforts. Using CNN and transfer learning (Hassan et al. 2020) analyzed the visual sentiment from disaster images in social media, which can estimate sentiments such as joy, fear, anger, sadness etc. from the images related to disasters. (Hao and Wang 2020) used two modules in their pipeline to analyze multimodal social media data, such as text and images. The designed image analysis submodule used five machine learning classifiers (support vector machine, logistic regression, artificial neural network etc) to estimate whether a hazard was present in the image, hazard type, hazard severity etc. Combining the text analysis they can provide a good description of disaster damage both from text and image. There are many other research efforts in this area, and many are building a pipeline for multi-hazards or multimodal datasets, which differ from the research we are going to show below.

In this paper, we use a transfer learning based approach to select images from social media platforms that contain damaged buildings, so they can be used in an application we developed in the earthquake science domain. The motivation of this work came from the earthquake crowdsourcing application MyShake (Qingkai Kong et al. 2016; Allen, Kong, and Martin-Short 2019; Qingkai Kong, Allen, and Schreier 2016; Q. Kong, Patel, and Inbal 2019; Qingkai Kong, Martin‑Short, and Allen 2020), which recently started to allow users to upload felt reports after an earthquake and display the felt region to the users (Strauss et al. 2020; Rochford et al. 2018). These reports are short surveys that the users can complete with pre-defined options for the felt experience and observation of damaged structures. With these felt reports, there are also plans to enable image uploading from users after an earthquake, which may include images of damaged buildings, roads, and bridges in the area. These images are not only useful to the MyShake users to learn where there is damage around them, but also can provide a valuable dataset for the research community and emergency services communities, especially coupled with locations (because felt reports in MyShake also request the user location). Potential applications can be developed from this information including the severity of the damage, better maps of shaking distribution and damage, and prioritization guidance for the rescue community. But users can upload any images, including images not related to damage, therefore, we must build a machine learning based approach to select only the images containing damaged buildings.

The approach and results described below include a model we built by using transfer learning with a pre-trained model VGG19 (Simonyan and Zisserman 2014), which is a model trained on the ImageNet dataset (Deng et al. 2009). Transfer learning has been shown to be a very effective way to build a model in many different fields (Weiss, Khoshgoftaar, and Wang 2016) when obtaining a large target training datasets is a challenge. Instead, by fine tuning the last few layers of a well pre-trained model, good performance can be achieved. Using 6,556 manually labelled images downloaded from Twitter and Getty Images, we fine tuned the block 5 of the VGG19 model, and trained the model to recognize which image contains damaged buildings in the scene. To test the performance of the built model, we downloaded a new test dataset from Twitter during the period 2020-01-01 to 2020-09-30 with 20,804 images. The results from these tests are good, the damage building recall is 80.5% with a precision of 77.4%; non-damage building recall is 99.6% with a precision of 99.6%. In addition to evaluating the model using the test dataset, we also ran the model in near real-time for 30 hours on Twitter after the 2020-10-30 M7.0

Turkey earthquake. The results for the real-time test on the Turkey earthquake are also good, with a damage recall of 77.7%, and damage precision of 94.7%. Non-damage recall was 94.4%, and non-damage precision was 98.9%. To understand what the model learns, we also implemented a Gradient-based localization method, Grad-CAM (Selvaraju et al. 2017), which can show us which part of the image the model is using to make the decision. By generating heatmaps using Grad-CAM, we can see the model indeed learned to find the damaged texture in the image to make the decision about damaged buildings.

## Results

### Performance of the transfer learning model

The trained model (see the method session for the details of the transfer learning) performs well on the reserved validation dataset, which consists of 533 damage building images and 717 non-damage building images (we call it the "other" category). Figure 1 shows the precision, recall and f-measure for both classes by using different thresholds on the validation dataset, and we decided to use 0.5 as our final threshold. This gives us precision and recall for the damaged building class of 0.89 and 0.87 respectively, while the precision and recall for non-damage building class are 0.91 and 0.92. Note the imbalanced nature of the dataset, i.e. there are more non-damage building classes than the damage building class. This will be especially true in real-world applications, therefore, we keep these precision and recall metrics calculated using the imbalanced dataset.

In order to test the model performance, we downloaded a new test dataset from Twitter in the time period 2020-01-01 to 2020-09-03, which includes 24,058 non-damaged images and 372 damaged images that are manually labelled by the authors. See table 1 of the confusion matrix for this test dataset, the corresponding damage building recall is 80.5% and precision is 77.4%; non-damage building recall 99.6% and precision is 99.6%. If we take a closer look at the images that are wrongly labelled, we can see where improvement is needed. Of the 73 damaged building images that the model missed, three types of mistake are common (see fig. 2 upper row): buildings with serious cracks (9 images), night views of the building (13 images), and the entire building ground floor collapsed (also known as soft story collapse) (18 images). These 3 types of errors account for 54.8% of all errors. Within the 88 images that the model mis-labelled as damaged buildings when they were not damaged, 73.9% come from three categories: 37 of them are maps with the majority of them from Google satellite images, 13 of them are aerial images that are taken from a far distance, and 15 images have features very similar to building debris, such as rock piles, tree leaves etc. The bottom row in Fig 2. gives examples of these images.

We did further tests of the model performance during the 2020-1030 M7.0 Turkey earthquake (USGS event id: us7000c7y0) in near real-time. We ran the model to monitor the incoming Twitter images for about 30 hours searching images with keywords "earthquake", and let the model classify these images into different categories. We then remove the duplicates of the images, and manually label these images to compare the results with the automatically generated labels. The confusion matrix for the output results is shown in Table 2. The model performs well with a damage recall of 0.78, damage precision of 0.95, non-damage recall of 0.99, and non-damage precision of 0.94.

Furthermore, we trained a 4-layer convolutional neural network (CNN) from scratch to serve as a baseline model for comparison purposes. The details of the baseline model is in supplementary material. But the performance of the baseline model is worse than that of the transfer learning. Using a threshold of 0.2, the damage building precision is 40%, while the recall is 86.0%. If a threshold of 0.5 is used, then the damage building precision is 51.1%, but the recall is 66.4%. Neither of these performances are close to the transfer learning performance.

**Fig. 1: The performance of the trained model on the validation dataset.**

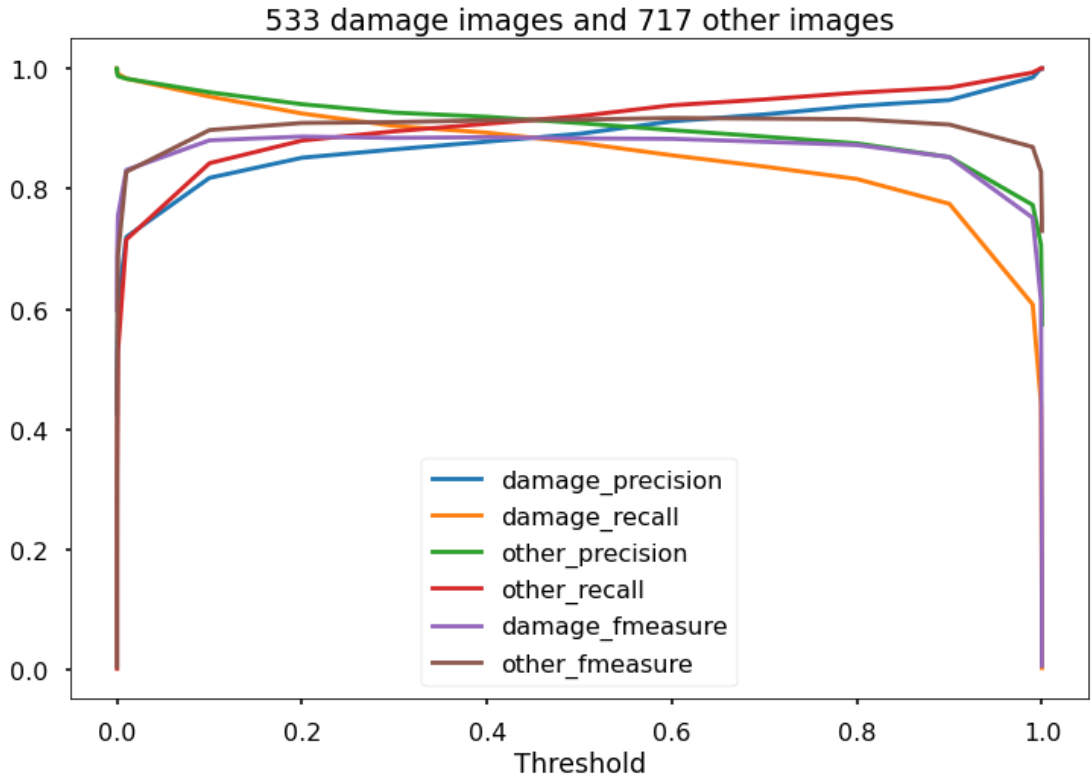

**Fig. 2: Examples of images that cause the trained model to commit a mistake.** The upper row shows examples of damaged buildings erroneously classified as non-damage. From left to right: cracks in the building, night view of the building, and the whole building yielded or soft story collapse. The bottom row shows the non-damage building images that were estimated as damaged building images. From left to right: Google satellite images, aerial view of the city and rock rubble similar to debris patterns.

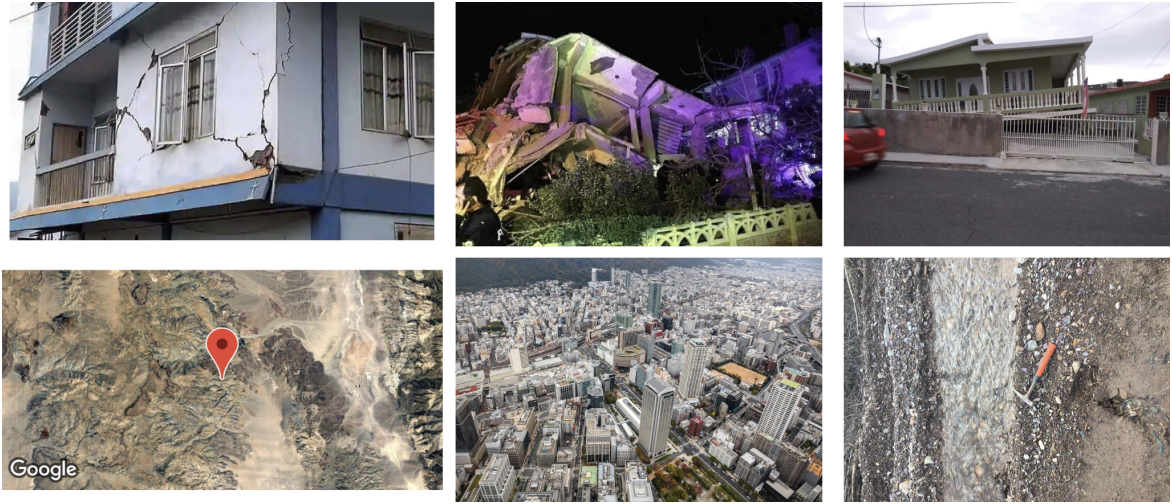

## Visualizing the results using Grad-CAM

We implemented the gradient-based localization algorithm - Grad-CAM to help us understand what aspects of the images the model was using to make the decision, the detailed description of the method is described in the method section. Fig 3 - 6 show the results of overlaying the Grad-CAM heatmap on the origin image for both classes estimations.

We show the heatmaps for both classes to have a better view of what the model was focusing on to make the decision. For example, fig. 3 top left panel, shows building damages. The left image in the panel shows where the model focuses if it thinks the image is a non-damage building class. Clearly, the model focuses on the sky part of the image, which is correct. The other panel shows where the model focused when looking for evidence of building damage. Overall, the probability that this image is a non-damage building class is 0.01, which means the model made a correct decision that this is a damaged building. There is a tendency that the more features of the damaged buildings that are in the image, the higher the confidence it becomes. The model learns to identify the debris of the buildings very well; debris piles are often the feature that is the focus of the model in the images. However, other features that are similar to the debris, such as grass fields and tree leaves, are occasionally mistreated as damaged debris. Examples of this are in figures 2 and 6. Figure 4 shows the true negatives, which the model correctly recognizes the images without damaged buildings. From the focused features in these images, we can see that, most of the time, the trained model focuses on the correct features to make the decision.

Figure 5 and 6 show the false negative and false positive cases, and we can see where the model focused when making the wrong decision. When there are damaged buildings, but the building is not the main focus of the image, then the image is often classified as non-damages building. This occurs when there are other objects in the foreground, or other objects occupy a large fraction of the image, or the buildings are in the dark (fig. 5). Also, less catastrophic damage such as cracks in the buildings do not lead to a damaged-building label. This is likely due to the fact that only severely damaged buildings are captured by the model, as the majority of the training data contains only the severely damaged buildings. Figure 6 clearly shows us where the model tends to make mistakes, i.e. treat the non-damage features as damage. As mentioned earlier, features that are similar to the debris, usually are determined as damage features by the model.

**Fig. 3: Grad-CAM heatmap on images that the model made the correct decision for damage-building class (True Positives).** Heatmap overlaying images indicates where the model pays more attention to make the decision; warmer colors indicate stronger influence on the decision. Each image is shown twice and overlaid with a heatmap indicating the regions of the image that suggest building damage, and the regions that suggest no building damage. The ground truth, estimated label and estimated probability is shown in the title above each image.

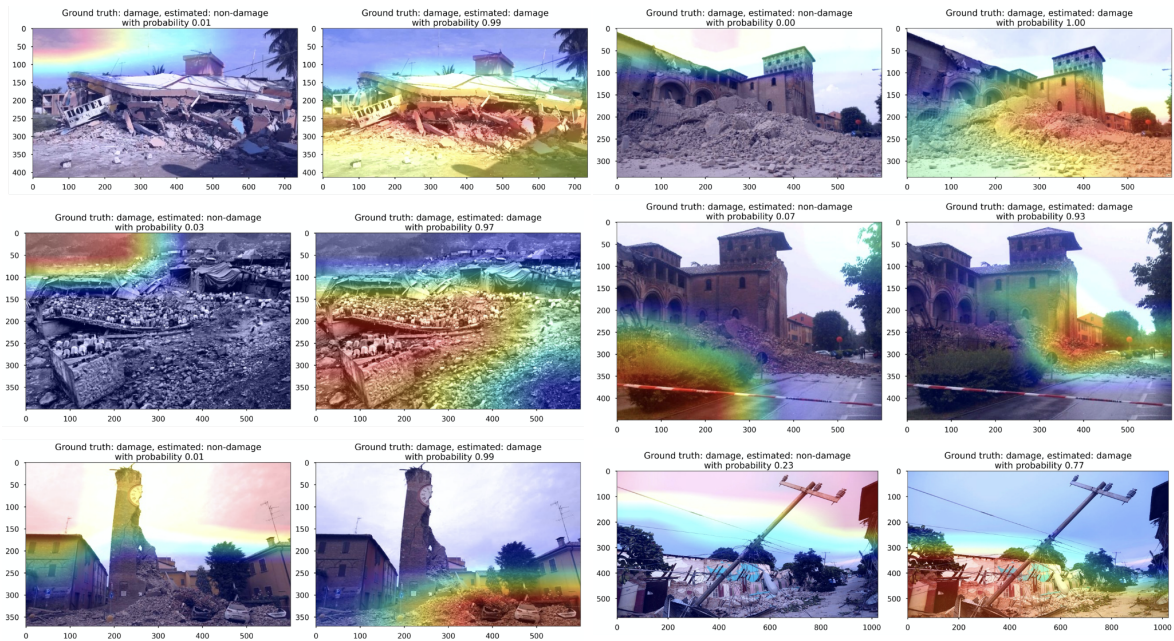

**Fig. 4: Grad-CAM heatmap on images that the model made the correct decision for non-damage building class (True Negatives).** Same as fig. 2 but with true negatives regarding the non-damage building class.

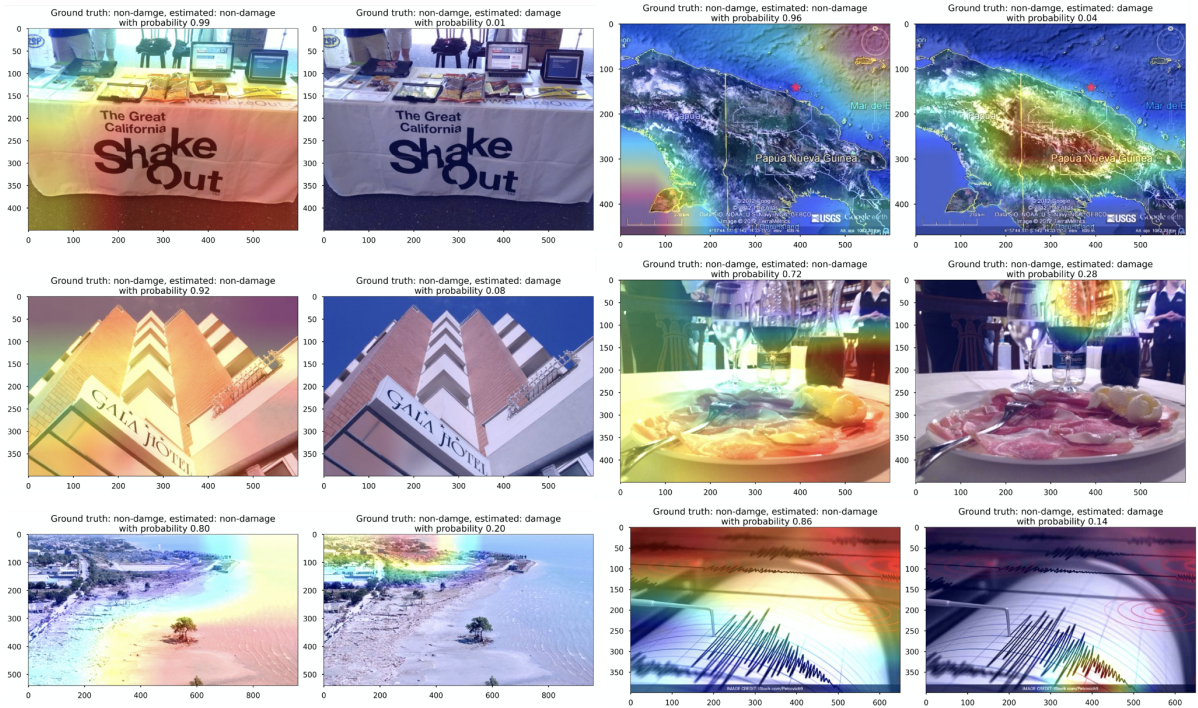

**Fig. 5: Grad-CAM heatmap on images that the model made the wrong decision for damage-building class (False Negatives).** Same as fig. 2 but with false negatives regarding the damage-building class. To protect the privacy of persons in the picture, we blockout their faces in the images.

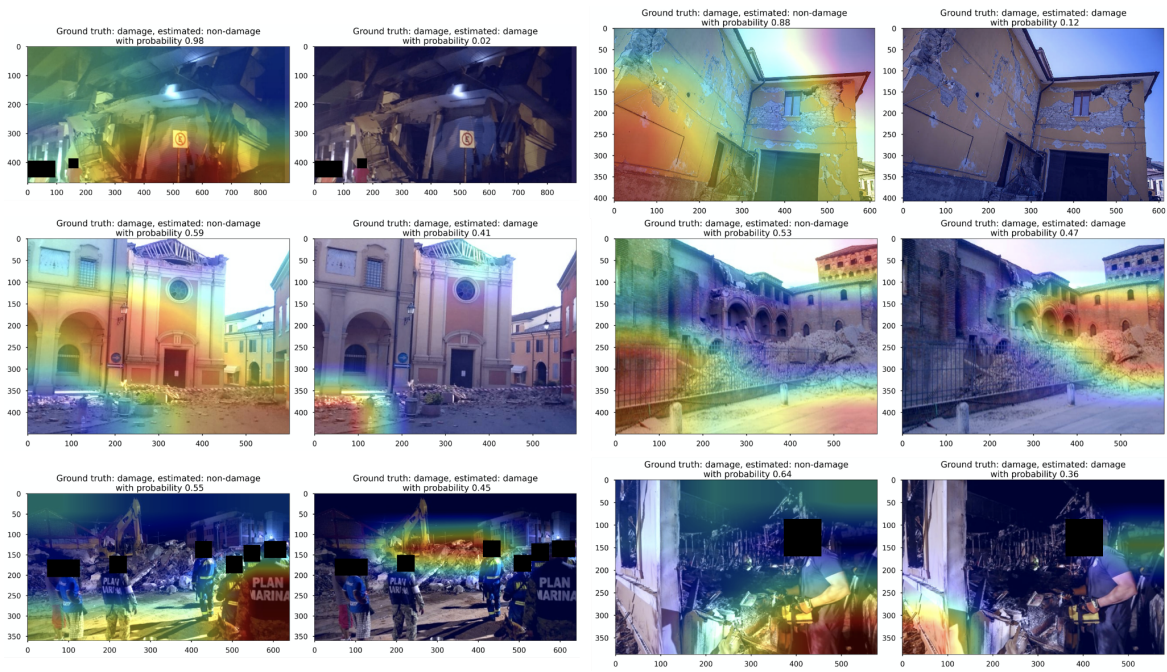

**Fig. 6: Grad-CAM heatmap on images that the model made the wrong decision for non-damage building class (False Positives).** Same as fig. 2 but with false positives regarding the non-damage building class. To protect the privacy of persons in the picture, we blockout their faces in the images.

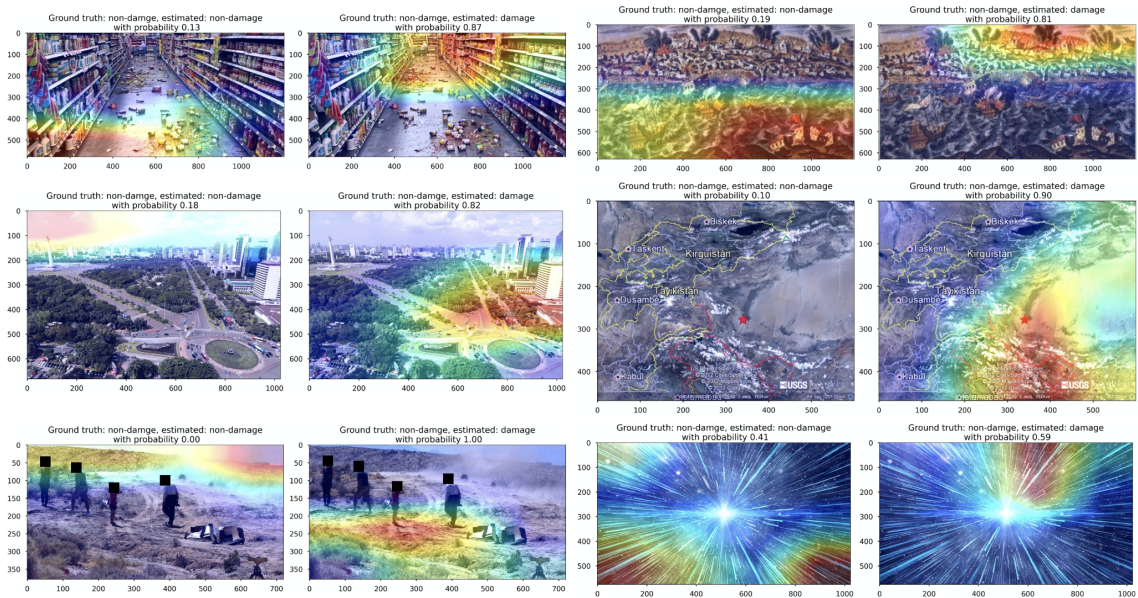

## Discussion

We designed this transfer learning approach with a more specific application in mind. We want to use it in the MyShake app to identify the images with damaged buildings that users may upload in the future after an earthquake. But the same method can be expanded broadly to extract images from social media platforms after a disaster. The implemented Grad-CAM method can help us to understand the learned features of the model and also diagnose potential improvements.

There are three main benefits of the pipeline we built here. Firstly, good performance can be obtained from a relatively small labelled dataset, thanks to the success of transfer learning developed in recent years. This for the many applications where collecting large amounts of labelled data is time consuming and costly. Here, we only used a few thousand labelled images to obtain a good working initial model. In addition, the computation will be dramatically reduced using transfer learning compared to building a model from scratch. Training is taking about 20s for each epoch on one K80 GPU, and with less than 20 epochs the transfer learning can converge to a good solution due to the smaller sets of trainable parameters. While training a whole baseline CNN model from scratch needs about ~1.5h to converge on the same GPU with a much simpler structure. It is also notable that the performance of the baseline model trained from scratch on this small dataset is much inferior to that of the transfer learning model.

Secondly, the Grad-CAM method also shows whether the trained model focuses on what we expect. We can see in Grad-CAM overlaid images that the current trained model can focus on the obvious damage of the building correctly. However, it focuses on the features from severely damaged buildings, such as the debris. From the false positives we can see the model also confuses patterns like debris, such as tree leaves, grass fields etc with the damaged building features. We find 73.9% of images in the false positives are due to the features in the images that look like debris, either from aerial images, satellite maps or tree leaves. This is an area that we can focus on to make improvements in the future by providing more training data that contain this type of patterns.

Finally, while the current model is not perfect, it is a good initial model that we can use to filter a large number of images online. The challenge we face with this problem is that there are many more images of undamaged buildings than damaged. We can use this model to initially filter a much larger dataset of images, and then manually filtering out the false positives to further improve the model.

As we can see from the performance section, the model can be improved in a few places. First of all, the false positives of the damaged buildings mainly come from the patterns that are similar to the debris, such as the rock rubbles and tree leaves. In addition, satellite maps and aerial images can also trick the model, as some of these textures look like damaged building materials. One potential solution for this problem is to collect more images of these kinds and add them into the training dataset to make the model more robust for these features. One good aspect is that these types of images are abundant on social media platforms and easy to collect compared with the damaged building images, therefore, this could be the next step to improve the model.

Secondly, for the false negatives of the damaged buildings, such as the night scenes, and cracks in buildings, may need a different approach other than collecting more training data, since these types of images may not be common on the social media platforms. For the night scene images, due to the low light source during the time of taking the pictures, the building objects in these images are usually blurry and dim, making it intrinsically hard to estimate the class. One way to get more of these night time images with damaged buildings is to artificially convert the day time images into night time images using image processing packages. This way, we can turn the images taken in day time after the earthquakes to night time, essentially a good data argumentation step. On the other hand, there are also many research developments of enhancing the night scene images for object detection or even converting the night scene images to the day time images (Priyanka, Wang, and Huang 2019; Priyanka, Tung, and Wang 2016; Capece, Erra, and Scolamiero 2017; Ai and Kwon 2020). This approach could potentially make these night scene images easier for the model to assess. As for the cracks in the buildings, this is perhaps not too much of a problem. The purpose of this project is to detect severely damaged buildings after the earthquake. If we want to capture these cracks related to smaller earthquakes in the future, we can reach out to the civil engineering community for more training data as they are also collecting and developing machine learning models to identify building cracks after the disaster (Li and Zhao 2019; Özgenel and Sorguç 2018; Hoang 2018; Eschmann et al. n.d.). The soft story building collapse is another type of failure for which we need to make improvements. One way is to do more data argumentation (such as shift, rotate, increase contrast, brightness etc) on these types of images to increase the number of training cases in the current training datasets. But a better way is to reach out to the civil engineering community for more training image data.

As stated in the motivation of this study, we hope that in the future MyShake users can upload images they take after an earthquake to report building damage. These images could be displayed on the map in the app to better inform users about damage in their neighborhood. As the databases of these images are accumulated overtime, we could also assign a score for the severity of the damage in each image. At the same time, this collected database of damaged structure images could be used by the civil engineering community to quantify the damages after an earthquake.

## Methods

## Transfer learning with VGG19

We built our model on top of the VGG19 model, which was trained on the ImageNet dataset. Figure 7 shows the transfer learning architecture. The first 4 blocks of layers were locked and we only allowed the weights in the 4 convolutional layers in block 5 to vary to fine tune the model. After flattening the feature maps from block 5, a fully connected layer with 265 neurons was added before feeding into the softmax layer to make the final decision.

Input images were rescaled to 150 pixel by 150 pixel for all 3 channels and we normalized the values between 0 and 1. Due to the imbalance of the training datasets (3684 non-damage images and 2872 damage building images), we use the class weights 0.8897 for non-damaged class and 1.1414 for damaged building class to weight the loss function (during training only) to tell the model to "pay more attention" to samples from damage building class. A categorical crossentropy loss function was used with the RMSprop optimization algorithm (with initial learning rate set at 1e-5). The training accuracy curve is shown in Figure 8, and training was conducted on one K80 GPU using Google Cloud Platform with each epoch taking about 20s. The best model was achieved at epoch 8. For more details, please see the python notebook on the provided github repository (see the Code Availability section).

**Fig.7: Structure of the transfer learning model.** The input is (150, 150, 3) meaning the image dimension is 150 by 150 pixels, with 3 channels. Convolutional layers are the yellow boxes with kernel size written in the front and number of kernels in the end, for example 3 x 3 conv 128 means kernel size is 3 by 3 pixels with a total of 128 kernels. Blue boxes are max pooling layers. Green box represents the fully connected dense layer with 256 neurons. Red box is the output layer with the softmax layer to output the classification results for the two classes.

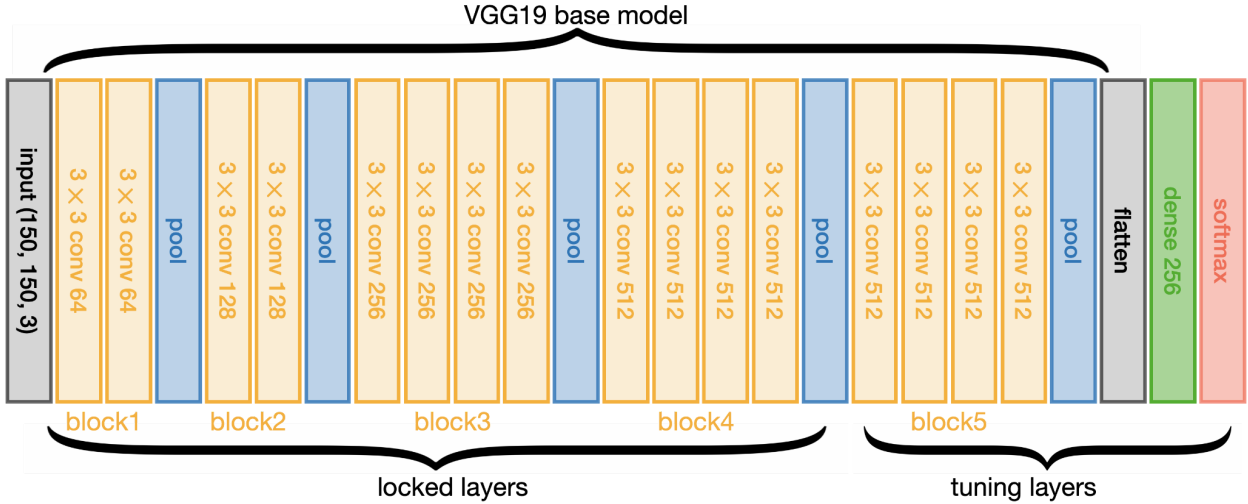

**Fig.8: Training accuracy curve.**

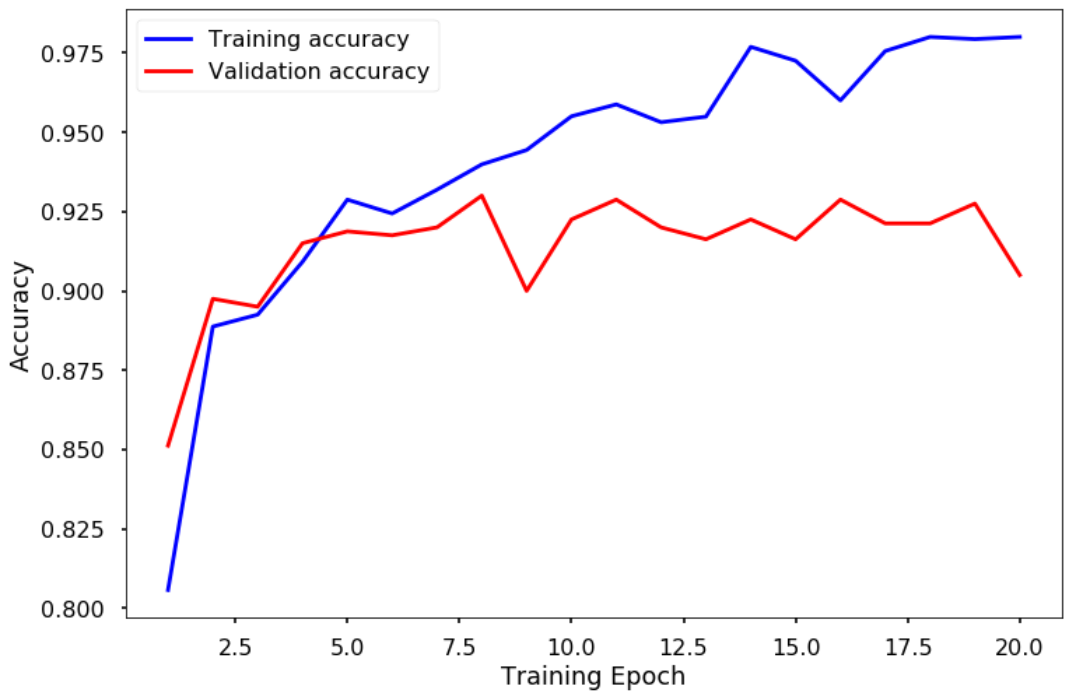

## Baseline CNN model

In order to have a baseline performance, we also built a standard convolutional neural network to serve as a baseline model. Due to the relatively small labeled dataset, we only used up to 4 CNN layers in the model. The model structure and training details can be found in the supplementary material.

## Grad-CAM

To better understand what the model is relying on during the inference, we visualize the deep neural network we built via the gradient-based localization algorithm - Grad-CAM (Selvaraju et al. 2017). The basic idea of this method is that the last convolutional layer extracts the feature maps that have the best compromise between high-level semantics and detailed spatial information, and that the whole model will make the decision based on these feature maps. By taking the gradients of the class score with respect to the feature maps, they provide a good indication of the pixels that are important to the final decision.

More specifically, using the symbols used in the Grad-CAM paper (Selvaraju et al. 2017), we can first take the gradient of the score $y^c$ for class c, with respect to the last convolutional layer feature maps $A^k$, where k is the $k^{th}$ feature map, and globally average them across all the pixels to obtain the importance weights $\alpha_k^c$ for each feature map, as shown in the equation (1):

$$\alpha_k^c = \frac{1}{Z}\sum_i^u \sum_j^v \frac{\partial y^c}{\partial A_{ij}^k} \quad (1)$$

where u and v are the width and height of the feature map, and i and j are the (i, j) location on the feature map. These weights give the relative importance of each of the feature maps for the target class c before the model makes decisions. Then the class discriminative localization map Grad-CAM $L_{Grad-CAM}^c$ can be calculated using equation (2):

$$L_{Grad-CAM}^c = ReLU(\sum_k^n \alpha_k^c A^k) \quad (2)$$

where n is the total number of feature maps in the last convolutional layer, and ReLU is the Rectified Linear Unit. The reason to apply a ReLU to the linear combination of the weighted feature maps is that only the features that have a positive influence on the target class are needed. The derived localization map is a non-negative weighted average of all the feature maps in the same dimension of the last feature map, and it is up-sampled to the input image resolution using bi-linear interpolation to form the final heatmap. This heatmap can show the importance of the different regions on the input image influencing the final output to the target class.

## Data Collection and Availability

All the data described below are listed as the image ids in the csv files in the supplementary materials. Please note that due to deletion from the owners of the original post/image, a small portion of the links may not work. All the images shown in this paper are from Twitter which are subject to the policy of free use in Non-commercial usage: https://developer.twitter.com/en/developer-terms/agreement-and-policy. The Getty images used in this paper are only downloaded from the website for training purposes, and have been deleted after training. This follows the requirements from Getty Images "You are welcome to use content from the Getty Images site on a complimentary basis for test or sample (composite or comp) use only, for up to 30 days following download. However, unless a license is purchased, content cannot be used in any final materials or any publicly available materials. No other rights or warranties are granted for comp use." from https://www.gettyimages.com/eula.

**Training Dataset**

For this work, we decided to use training images from earthquake events at two locations: Nepal and Indonesia. We then test the trained model on other locations/events to assess its performance. We began the data collection with Twitter as the source since high social media activity occurs on this platform following earthquakes and such activity on Twitter is predominantly open for public access.

A historical search of the Twitter feed was conducted for the 2015 Nepal earthquake event. An advanced search was conducted using GetOldTweets3 python library (https://pypi.org/project/GetOldTweets3/) and `tweet_id` were extracted from the results. The following search parameters were used: Keyword: earthquake; Filter: images; Date Range: April 25, 2015 to April 30, 2015.

Subsequently, the Twython python library (https://twython.readthedocs.io/en/latest/) was used to fetch the url of each image attached to every `tweet_id`. It is quite frequent on Twitter that many tweets can reference the same image. Thus, we only fetched images with unique urls. A total of 47785 images were obtained from this search, which contained pictures of damaged buildings, other damaged structures (e.g. roads, bridges, etc.), and a wide variety of other pictures related and unrelated to the earthquake event. A number of cleaning operations were performed on this dataset are described as follows: firstly, we found that even though we fetched images with only unique urls, many of them were still duplicates, which was assessed by comparing md5 hash of the images. All such duplicates were removed from the dataset.

Secondly, we chose to remove images that were smaller than 150px in either height or width. Thirdly, we found that there were many images that were even though not exact duplicates, they represented the same shot but differed in either just pan or zoom. Such images were processed through a proprietary software, Duplicate Photo Cleaner. This software allowed us to identify images that are similar by specifying a user chosen extent of similarity (in percentage). In this study, we chose a threshold of 50% similarity and only 1 copy from each set of images identified as "similar" was retained and the rest were removed. The total number of images thus obtained from Twitter post these operations was reduced from 47785 to 18238. A manual operation was performed to label a subset of the obtained images in two classes, viz. 'damage' and 'non-damage'.

An image was labelled 'damaged' when more than ~20% of the image area contained visible damage of a building structure in the form of debris, rubble, cracks, etc. See the upper 8 images in Figure 9 for examples. The 'non-damage' label was applied to any image which either contained buildings that were not damaged and/or contained completely unrelated pictures, like a portrait of a person, text, map, etc. See the lower panel 8 images in Figure 9 for a few examples.

**Fig.9: Training data examples.** The upper panel 8 images are labelled damaged buildings, while the lower panel 8 images are labelled non-damaged buildings. All images are from Twitter.

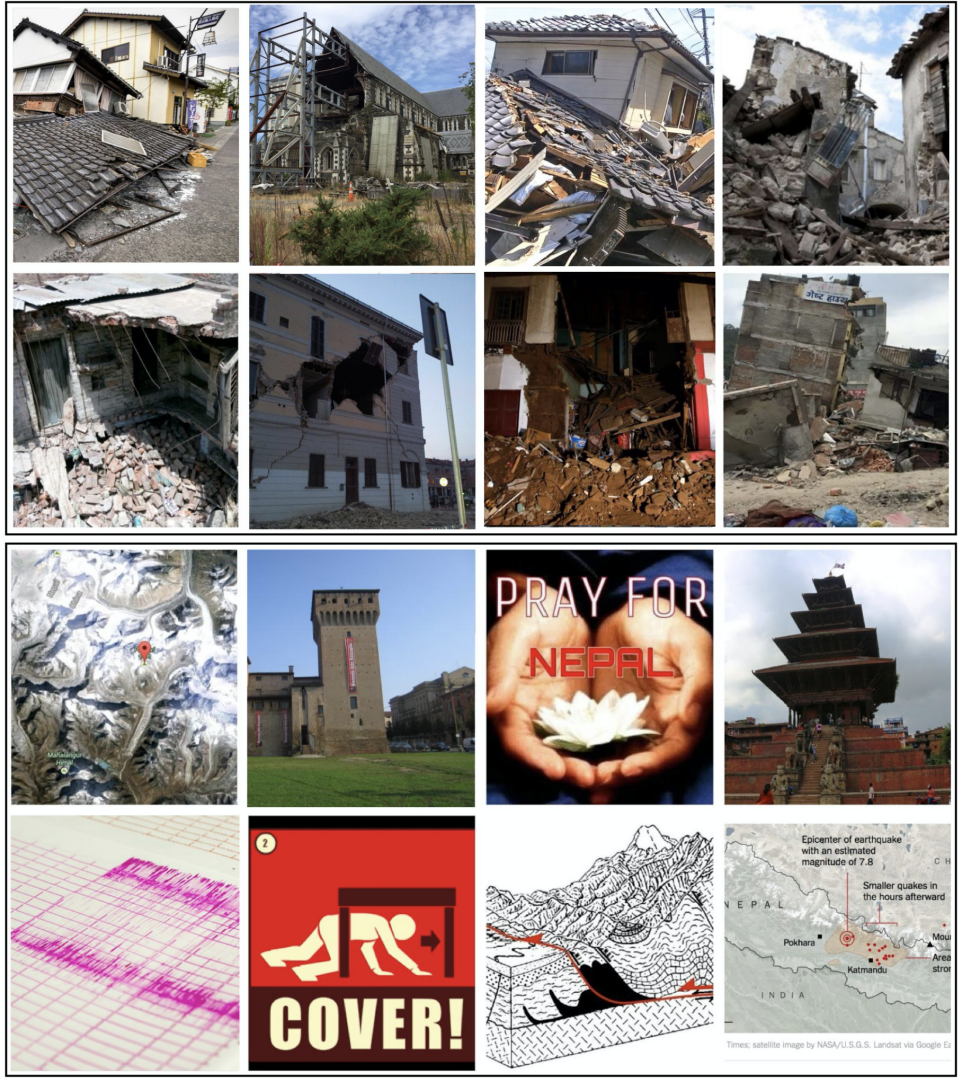

In all, 2337 images were manually labelled out of which 367 were classified as 'damage' while 1970 as 'non-damage' images. Thus, only about 16% images were obtained of the 'damage' class. In order to obtain images of 'damage' classification with better efficiency, we explored Getty Images as another data source. Publicly available images on Getty Images platform (https://www.gettyimages.com/) were obtained using an advanced search. We used "earthquake destruction" as our search term while setting the location to Nepal and Indonesia individually for these two places.

This resulted in 4219 images from Getty Images which were subsequently labelled manually. 2505 images were labelled as 'damage' class while 1714 images were labelled as 'non-damage' class. A csv file of the web links to these images made publicly available by Getty Images for free is included in Supplementary Material.

The two sets of data (from Twitter and Getty Images) were combined to form the training dataset of a total of 6556 images, comprising 2872 'damage' class and 3684 'non-damage' class. A schematic on this data collection process is shown in Figure S4 in the supplementary material. The links to the these images are stored in the Table s3 and s4.

**Excluded Images**

Certain images were found to not meet either of the criteria above and were excluded from the training dataset. Figure S5 and S6 in the supplementary material illustrates a few examples of such images. They essentially fall in two categories: (1) The damaged structure is not a building. Such images may be of damaged roads, vehicles, building interiors, etc. (2) The picture consists of elements other than the original shot. This included overlaying text on the image, screenshot of a news clipping, collage of different images, etc.

**Validation and Test Dataset**

Validation Dataset

The validation dataset was prepared with an aim to tune the threshold. These images were collected from Getty Images with "earthquake destruction" as our search term with locations set as "Italy" and "Mexico".

1250 images were obtained from Getty Images and subsequently were labelled manually. 533 images were labelled as 'damage' class while 717 images were labelled as 'non-damage' class. A schematic on this data collection process is shown in Fig. S7. Links to these images are stored in Table s5 and s6.

Test Dataset 1

A historical search of the Twitter feed was conducted for any earthquake related tweets between Jan 1, 2020 and September 30, 2020. An advanced search was conducted by employing the similar approach as for the training dataset, using the python libraries GetOldTweets3 (https://pypi.org/project/GetOldTweets3/) and Twython (https://twython.readthedocs.io/en/latest/). The following search parameters were used: Keyword: earthquake; Filter: images; Date Range: Jan 1, 2020 to September 30, 2020. A total 24640 images were obtained. After removing the duplications, we have 20804 images left. Links to these images are stored in Table s7 and s8.

Test Dataset 2

The second test dataset was prepared with the aim of running the model in near real-time immediately after a major earthquake event. During the course of this work, an earthquake occurred in Turkey on October 30, 2020 (2020-1030 M7.0 Turkey earthquake). A live stream of the twitter feed was analyzed to capture Twitter activity related to the earthquake. The live Twitter feed was processed for a total of 29 hours and 45 minutes between 10/30/2020 21:15:00 PST and 11/1/2020 20:22:00 PST using the keyword "earthquake" and filter "images". Due to some connectivity issues during the live stream the connection had to be reestablished a few times, thus reducing the total number of active hours of data retrieval. A total of 992 images were obtained after removing the duplicates. Links to these images are stored in Table s9 and s10.

## Code Availability

The code for the demo can be found in Github repository:

https://github.com/qingkaikong/damaged_building_detection.

## Acknowledgements

This work was initiated from the Data Science Discovery Program from Berkeley Computing, Data Science, and Society. We appreciate the opportunities this program provided to both researchers and students. In this program, students Gaurav, Srujay, and Jennifer played an important role to conduct this project and experience the whole process of doing research. This work is also done under the framework of MyShake. The Gordon and Betty Moore Foundation funded this analysis through grant GBMF5230 to UC Berkeley. The California Governor's Office of Emergency Services (Cal OES) funded this analysis through grant 6142-2018 to Berkeley Seismology Lab. We thank the previous and current MyShake team members: Roman Baumgaertner, Garner Lee, Arno Puder, Louis Schreier, Stephen Allen, Stephen Thompson, Jennifer Strauss, Kaylin Rochford, Akie Mejia, Doug Neuhauser, Stephane Zuzlewski, Asaf Inbal, Robert Martin Short, Sarina Patel and Jennifer Taggart for keeping this Project running and growing. All the analysis of this project is done in Python. We thank all the MyShake users who contribute to the project! We also thank USGS, especially the great community tool/application DYFI to enable this study. We also thank all the MyShake users who contribute to the project! We also thank Google Cloud Platform for providing the $5000 research credit to enable the computational aspect of this project. We also thank Twitter and Getty Images for making data public for research purposes. Qingkai Kong's work was performed under the auspices of the U.S. Department of Energy by Lawrence Livermore National Laboratory under Contract Number DE-AC52-07NA27344. Jim Huang's contributions to this publication are his and were not performed on behalf of AT&T. Any opinions, findings, conclusions, or recommendations expressed in this publication are those of the authors and do not necessarily reflect those of the supporting agencies or companies. This is LLNL Contribution Number LLNL-JRNL-826989. Gaurav Chachra's work was performed in his personal capacity. It has no relation to either his job responsibilities at Zymergen or any intellectual property of Zymergen that he had access to.**Ethics declarations**

None

Tables

**Table 1 Confusion matrix for the Twitter test dataset after remove duplicates**

|  | Predicted damage | Predicted non-damage | Precision | Recall |
| --- | --- | --- | --- | --- |
| True damage | 301 | 73 | 0.774 | 0.805 |
| True non-damage | 88 | 20342 | 0.996 | 0.996 |

Table 2 Confusion matrix for the near-real time test on M7.0 Turkey Earthquake after remove duplicates

|  | Predicted damage | Predicted non-damage | Precision | Recall |
|---|---|---|---|---|
| True damage | 160 | 46 | 0.947 | 0.777 |
| True non-damage | 9 | 777 | 0.989 | 0.944 |

# Supplementary Information for Detecting Damage Building Using Real-time Crowdsourced Images and Transfer Learning

Gaurav Chachra , Qingkai Kong, Jim Huang, Srujay Korlakunta, Jennifer Grannen, Alexander Robson, Richard Allen

# Supplementary Figures

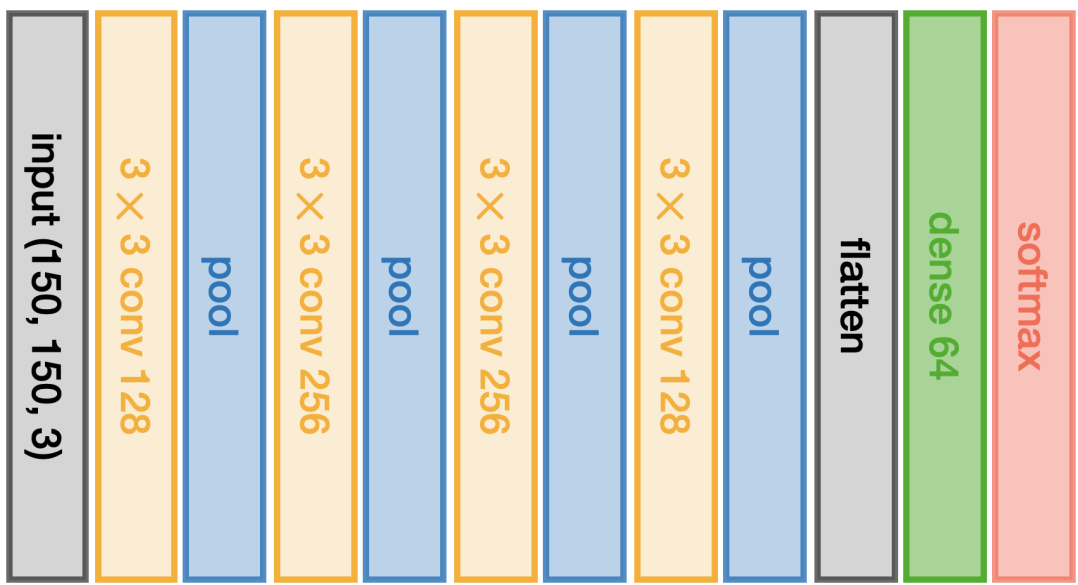

Fig. S1 - The structure of the baseline CNN model. Different colors of the blocks represent the different layers. The input is (150, 150, 3) meaning the image dimension is 150 by 150 pixels, with 3 channels. Convolutional layers are the yellow boxes with kernel size written in the front and number of kernels in the end, for example 3 x 3 conv 128 means kernel size is 3 by 3 pixels with a total of 128 kernels. Blue boxes are max pooling layers. Green box represents the fully connected dense layer with 64 neurons. Red box is the output layer with the softmax layer to output the classification results for the two classes.

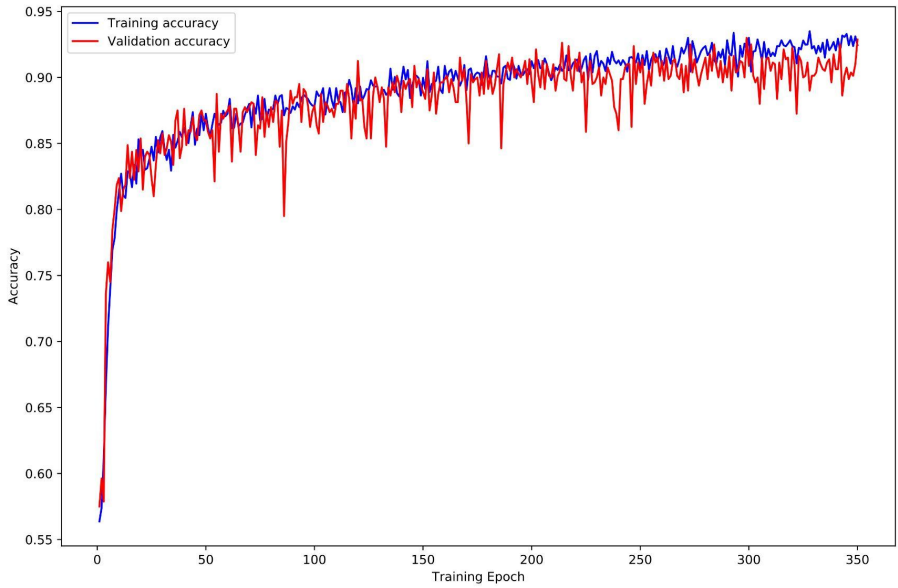

Fig. S2 - The training curve for the baseline CNN model.

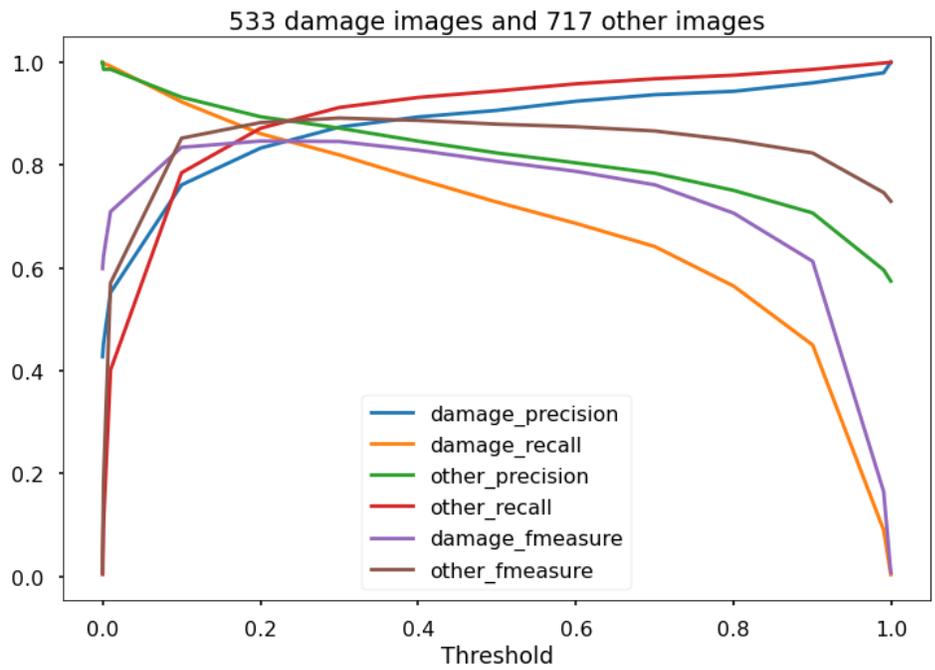

Fig. S3 - The validation dataset performance.

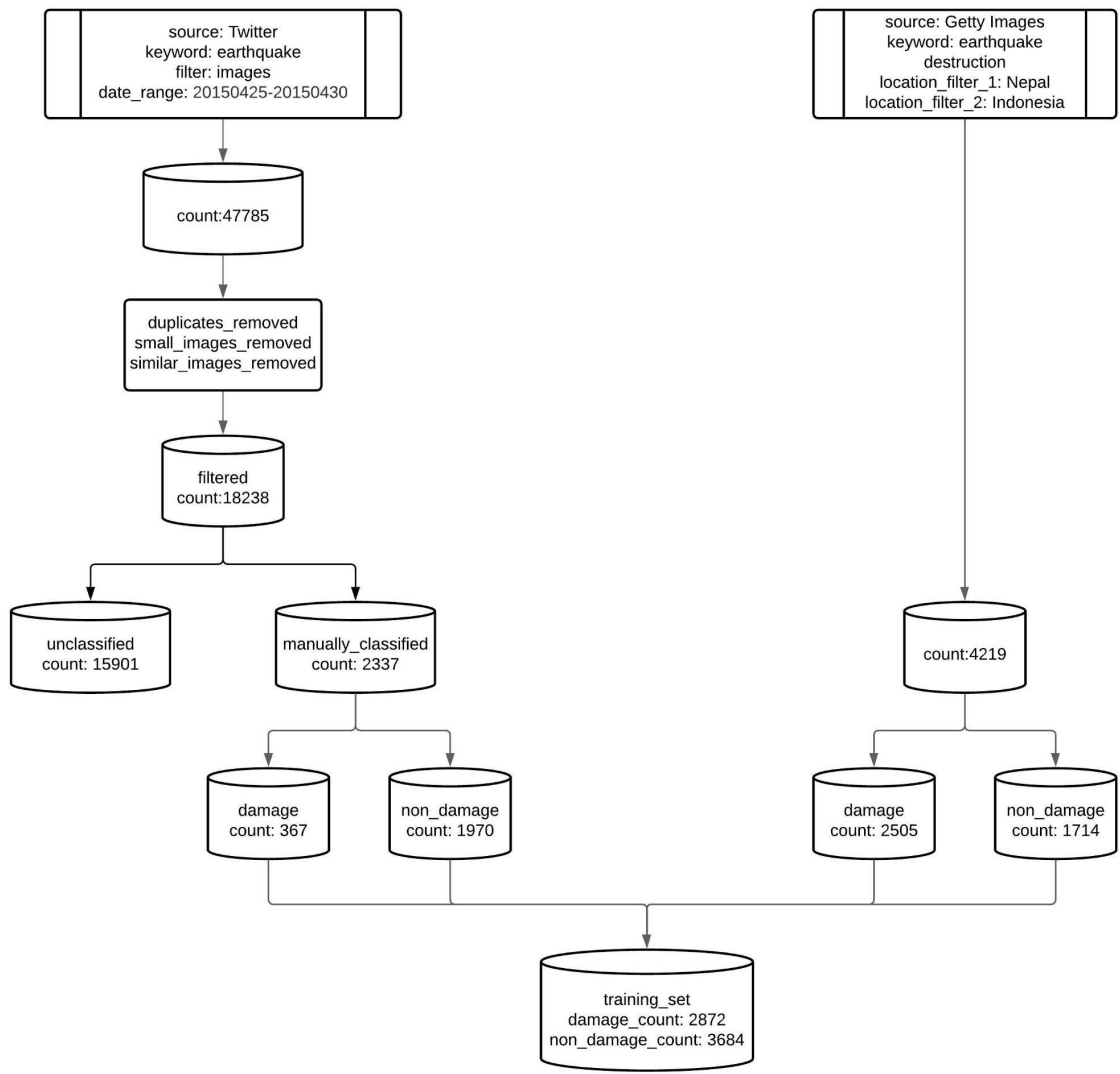

Fig. S4 - The schematic showing the process of collecting the training dataset.

### Damaged roads

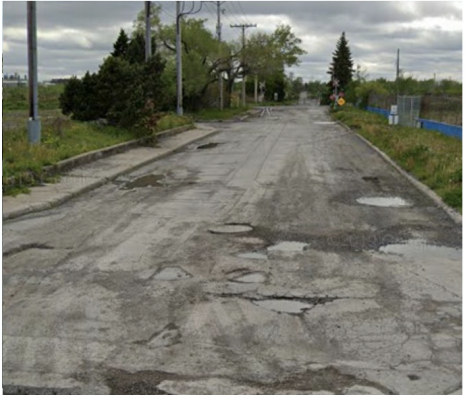 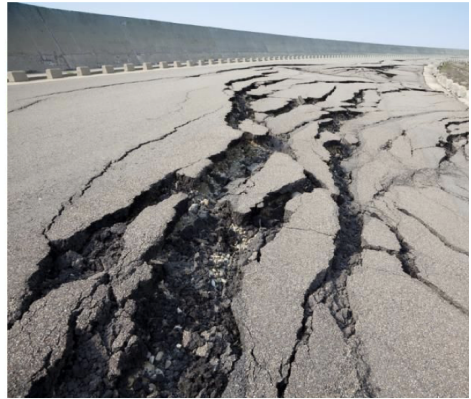

### Interiors of a building

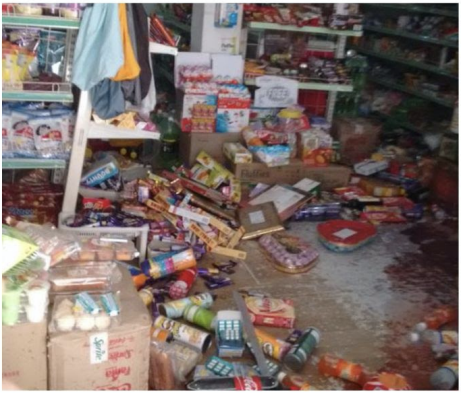 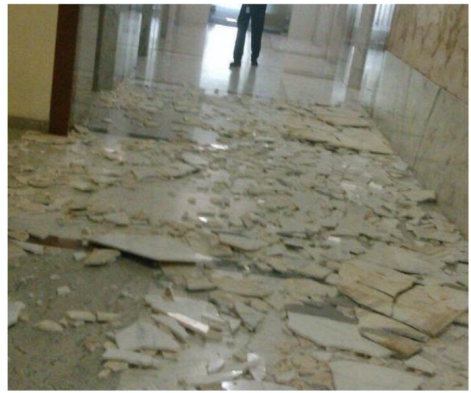

### Before/After shots

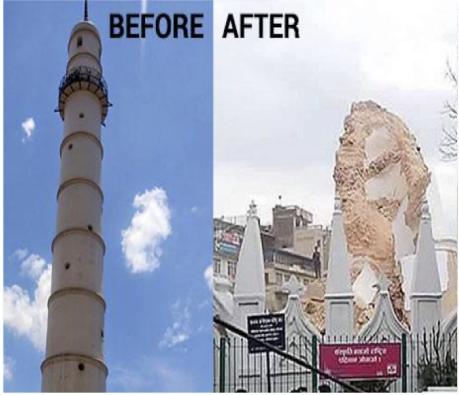 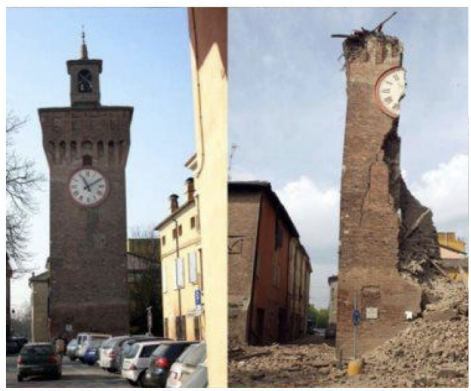

Fig. S5 - Images excluded from training dataset.

**News clipping/articles**

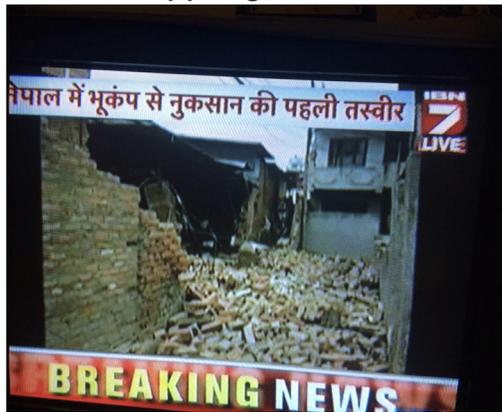
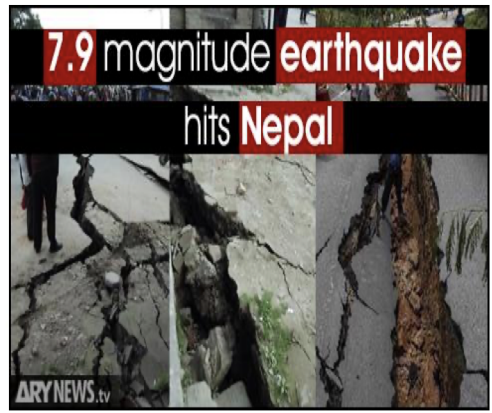

**Collage of different shots**

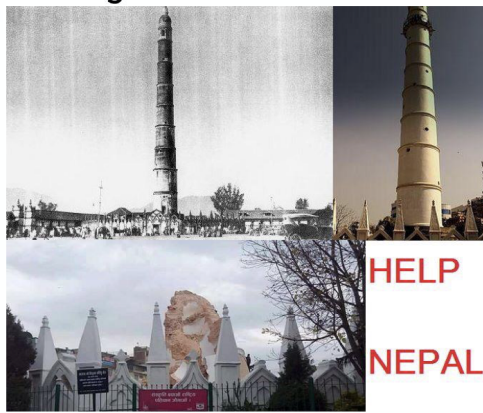
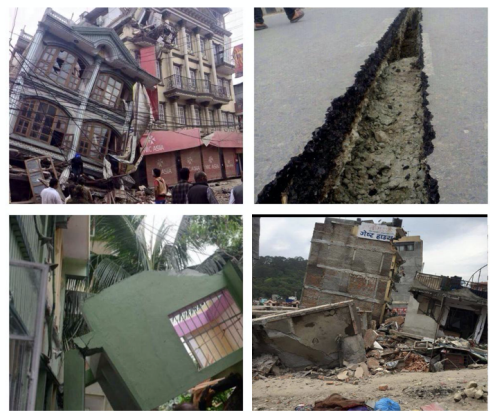

**Less than ~20% damage**

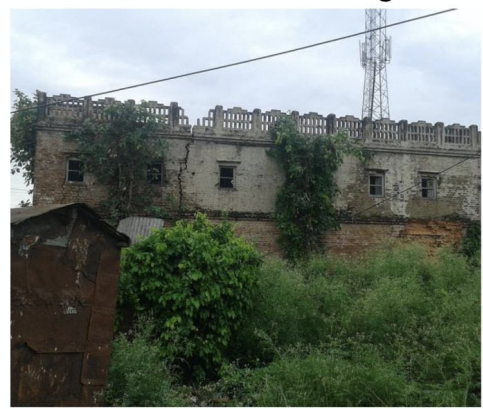
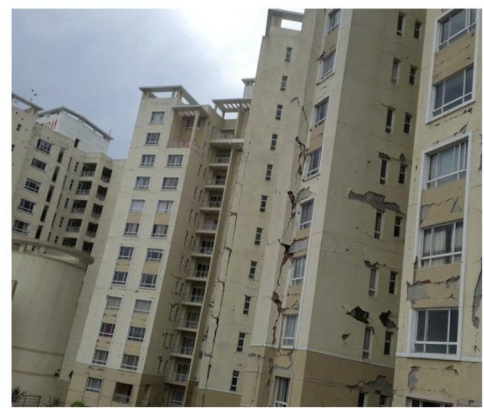

Fig. S6 - Images excluded from training dataset.

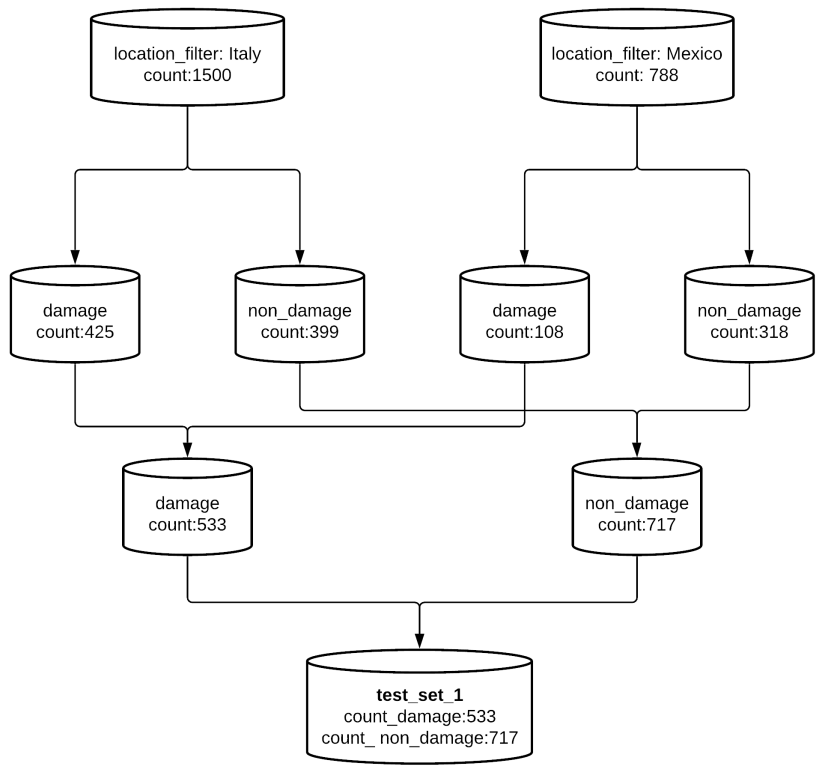

Fig. S7 - The schematic showing the process of collecting the validation dataset.

# Supplementary Tables

**Table s1 Confusion matrix for the baseline model using threshold 0.2**

|  | **Predicted damage** | **Predicted non-damage** |
|---|---|---|
| True damage | 320 | 52 |
| True non-damage | 472 | 23586 |

**Table s2 Confusion matrix for the baseline model using threshold 0.5**

|  | **Predicted damage** | **Predicted non-damage** |
|---|---|---|
| True damage | 247 | 125 |
| True non-damage | 236 | 23822 |

Table s3 - links to the damaged-class images used in training dataset.

Table s4 - links to the non-damaged-class images used in training dataset.

Table s5 - links to the damaged-class images used in validation dataset.

Table s6 - links to the non-damaged-class images used in validation dataset.

Table s7 - links to the damaged-class images used in test 1 dataset, i.e. from tweets between Jan 1, 2020 and September 30, 2020.

Table s8 - links to the non-damaged-class images used in test 1 dataset, i.e. from tweets between Jan 1, 2020 and September 30, 2020.

Table s9 - links to the damaged-class images used in test 2 dataset, i.e. from live Twitter feeds from 10/30/2020 21:15:00 PST to 11/1/2020 20:22:00 PST.

Table s10 - links to the non-damaged-class images used in test 2 dataset, i.e. from live Twitter feeds from 10/30/2020 21:15:00 PST to 11/1/2020 20:22:00 PST.